\documentclass[journal]{IEEEtran}

\usepackage{xcolor,soul,framed} 

\colorlet{shadecolor}{yellow}
\usepackage[pdftex]{graphicx}
\graphicspath{{../pdf/}{../jpeg/}}
\DeclareGraphicsExtensions{.pdf,.jpeg,.png}

\usepackage[cmex10]{amsmath}
\usepackage{array}
\usepackage{mdwmath}
\usepackage{mdwtab}
\usepackage{eqparbox}
\usepackage{url}

\usepackage[nocompress]{cite}
\usepackage{threeparttable}
\usepackage{graphicx}
\usepackage{amsmath}
\usepackage{amsfonts}
\usepackage{multirow}
\usepackage{multicol}
\usepackage{cite}
\usepackage[numbers]{natbib}
\usepackage{setspace}
\usepackage{color}
\usepackage{bbold}
\usepackage{bm}
\usepackage{makecell}

\usepackage{graphicx,amsmath,booktabs,latexsym}
\usepackage{amssymb}
\usepackage[lined,boxed,ruled]{algorithm2e}

\usepackage{subfigure}
\usepackage[switch]{lineno}
\usepackage{hhline}

\usepackage{cite}

\begin{document}
\bstctlcite{IEEEexample:BSTcontrol}
    \title{Conditional Goal-oriented Trajectory Prediction for Interacting Vehicles with Vectorized Representation}
    
  \author{
	Ding~Li$^{\footnotemark 1, 2}$,
	Qichao~Zhang$^{\footnotemark 2, 1}$,
	Shuai~Lu$^{\footnotemark 3}$, Yifeng~Pan$^{\footnotemark 3}$,
	Dongbin~Zhao$^{\footnotemark 2, 1}$,~\IEEEmembership{IEEE Fellow}

 \thanks{This work was supported by the National Natural Science Foundation of China (NSFC) under Grants No.62173325, and also by the Beijing Municipal Natural Science Foundation under Grants L191002, and in part by Science and Technology Innovation 2030 'New Generation Artificial Intelligence' Major Project No. 2020AAA0103700.}
}  
\maketitle
\footnotetext[1]{School of Artificial Intelligence, University of Chinese Academy of Sciences, Beijing 100049, China.}
\footnotetext[2]{State Key Laboratory of Management and Control for Complex Systems, Institute of Automation, Chinese Academy of Sciences, Beijing 100190, China (email: \{liding2020, zhangqichao2014, dongbin.zhao\}@ia.ac.cn).}

\footnotetext[3]{Baidu Inc., Beijing 100085, China (email: \{lushuai03, panyifeng\}@baidu.com).}

\markboth{IEEE TRANSACTIONS ON NEURAL NETWORKS AND LEARNING SYSTEMS, VOL. , NO. , 2021
}{Roberg \MakeLowercase{\textit{et al.}}: High-Efficiency Diode and Transistor Rectifiers}

\maketitle

\begin{abstract}
\textcolor{black}{Predicting joint future trajectories of two interacting vehicles faces great challenges, 
due to the high degree of multimodality and uncertainty in the future interaction process.} \textcolor{black}{For the traditional joint prediction models, each vehicle is treated as an individual agent, and its multimodal future modes can be learned from an efficient Goal-oriented Trajectory Prediction (GTP) method. However, such marginal methods produce multimodal self-consistent trajectories over individual agents without considering their future interactions.} \textcolor{black}{Motivated by this, this paper aims to tackle the interactive behavior prediction task, and proposes a novel Conditional Goal-oriented Trajectory Prediction (CGTP) framework to jointly generate scene-compliant trajectories of two interacting agents.} Our CGTP framework is an end-to-end and interpretable model, including three main stages: context encoding, goal interactive prediction and trajectory interactive prediction. First, a Goals-of-Interest Network (GoINet) is designed to extract the interactive features between agent-to-agent and agent-to-goals using a graph-based vectorized representation. Further, the Conditional Goal Prediction Network (CGPNet) focuses on goal interactive prediction via a combined form of marginal and conditional goal predictors. Finally, the Goal-oriented Trajectory Forecasting Network (GTFNet) is proposed to implement trajectory interactive prediction via the conditional goal-oriented predictors, with the predicted future states of the other interacting agent taken as inputs. In addition, a new goal interactive loss is developed to better learn the joint probability distribution over goal candidates 
between two interacting agents. \textcolor{black}{In the end, the proposed method is conducted on Argoverse motion forecasting dataset, In-house cut-in dataset, and Waymo open motion dataset.}
\textcolor{black}{The comparative results demonstrate the superior performance of our proposed CGTP model than the mainstream prediction methods.}
\end{abstract}


%
\IEEEpeerreviewmaketitle


\section{Introduction}\label{sec:introduction}

%
%
%
%
\begin{figure*}[htbp]
	\centering
	\begin{center}
		\scriptsize
		\includegraphics*[width=7.1in]{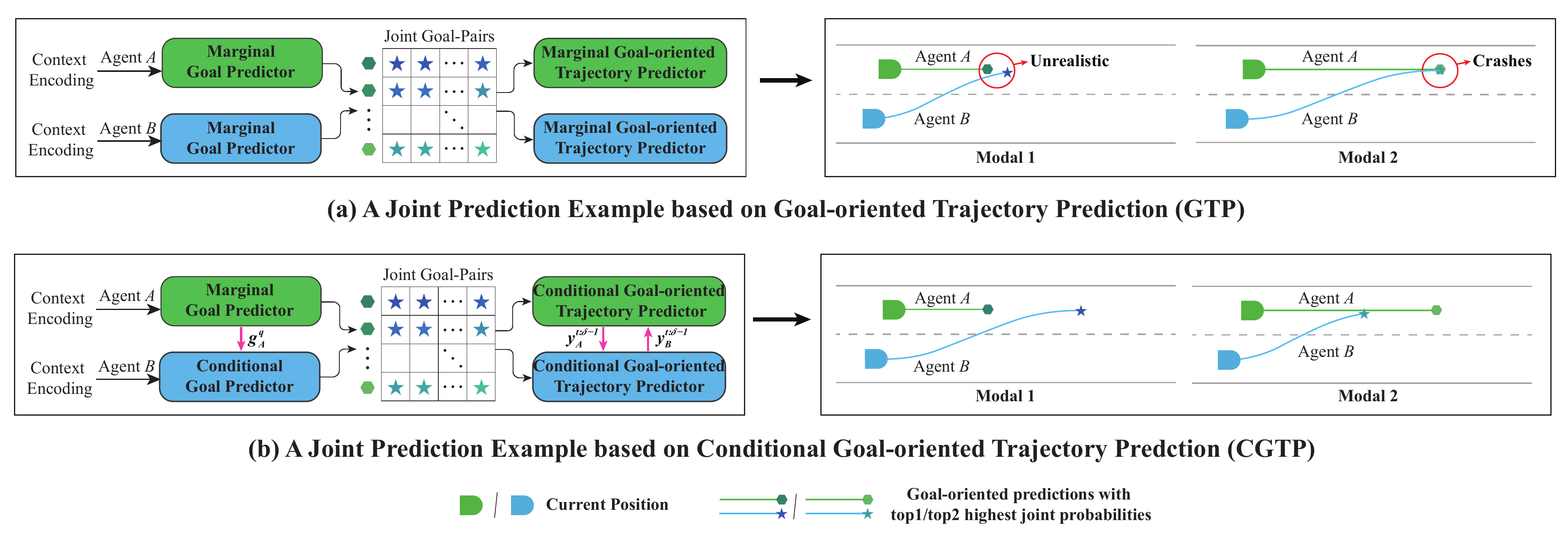}
		\caption{\textcolor{black}{\textbf{Motivation demonstration of the CGTP framework.}} 
		\textcolor{black}{In (a), the goal-oriented trajectory prediction model, a representative marginal prediction method, generates diverse goal-oriented future trajectories for each agent independently, producing joint predictions by combinations of self-consistent predictions over individual agents, with unrealistic or collision behaviors possibly occurring. In (b), the CGTP framework, a novel multimodal joint prediction method, accounts for future interactions via conditional modeling and outputs scene-compliant future trajectories. It focuses on goal interactive prediction realized by a combined form of marginal goal predictor with agent \textit{A} and conditional goal predictor with agent \textit{B}, and then the joint predictions can be generated from the conditional goal-oriented trajectory predictors.}}
		\label{fig:Motivation}
	\end{center}
   \vspace{-1.3em} 
\end{figure*}
Predicting or forecasting future trajectories of multiple agents is a mission-critical component in the autonomous driving system \cite{LKQ2021, DynaNet2021}, which plays an important role in the subsequent motion planning \cite{Brain2016, Lihaoran2020} and decision \cite{Wangjunjie1, Wangjunjie2} modules. In particular, the interactive behavior prediction has received increasing attention in recent years \cite{WaymoDataset2021}, which aims to jointly predict interactive behaviors 
of interacting agents in critical situations such as cut-in and yielding. 

\textcolor{black}{Fig.~\ref{fig:Motivation} presents two research pipelines focusing on the joint prediction fields. 
A naive approach for the joint prediction is to use a marginal prediction method. This class of models \cite{Desire2017, MultiPath2019, VectorNet2020, TNT2020} generate diverse predictions independently for each agent, and then make the combinations of marginal predictions to output the joint realizations. \textcolor{black}{Notably, the goal-oriented trajectory prediction (GTP) \cite{TNT2020} is a representative marginal prediction method, achieving great success in multimodal trajectory prediction by first identifying endpoints via the goal predictor and then predicting multiple trajectories via the goal-oriented trajectory predictor.} However, the weakness 
of such a method lies in the absence of future interactive modeling with the other interacting agent for the interactive behavior prediction. Limited by this, while the goal-oriented prediction method produces self-consistent predictions over individual agents, the joint prediction-pairs may result in unrealistic or collision behaviors. For instance, in spite of non-colliding predictions in the modal 1 of Fig.~\ref{fig:Motivation} (a), agent $B$ should aggressively speed up when making lane changing to keep a safe distance from agent $A$. Such scene-compliant interactive behaviors can be hardly captured by the pure marginal prediction method. To overcome this issue, recent advances have shown great success in predicting scene-compliant trajectories by learning from the multimodal joint prediction methods which adopt conditional prediction models to consider interactions between agent future predictions \cite{CBP2021,  mfp2019, Precog2019, ILVM2020}.}
\textcolor{black}{Especially, Waymo also provide a large-scale
interactive motion forecasting dataset \cite{WaymoDataset2021} for autonomous driving, and some multimodal joint prediction methods \cite{SceneTrans2021, sun2022m2i, ProspectNet} achieve better performances on this dataset.}
Hence, the multimodal joint prediction rather than marginal prediction is required for interactive driving scenarios.

\textcolor{black}{In the literature, the researchers characterize the underlying intents of interacting agents into various forms of future prediction information, which served as conditional dependencies to the multimodal joint prediction methods. 
A family of methods \cite{CBP2021,  mfp2019, Precog2019, SceneTrans2021, sun2022m2i, ProspectNet} leverages future overall trajectories of interacting agents as explicit future intents, focusing on interactive behavior prediction conditioned on them. Instead, ILVM \cite{ILVM2020} adopts implicit latent variables to describe the future intents over interacting agents. Different from recent studies above, in this paper, we propose a novel conditional goal-oriented trajectory prediction (CGTP) framework} \textcolor{black}{by combining significant advantages between the goal-oriented trajectory prediction method and conditional prediction method, as indicated in Fig.~{\ref{fig:Motivation}} (b). On one hand, CGTP conducts conditional inferences both in goal predictor and goal-oriented trajectory predictor by comparison with the goal-oriented trajectory prediction method. On the other hand, compared with the recent studies on multimodal joint prediction, CGTP focuses more on the future interactions at the goal-based level via a combined form of marginal and conditional goal predictors.} 
We factorize the CGTP framework into three parts: context encoding, goal interactive prediction and trajectory interactive prediction. To summarize, we list the main contributions of CGTP framework as follows:
\begin{itemize}
	\item For the context encoding, we design a Goals-of-Interest Network (GoINet) by combining the advantages of Graph Neural Network (GNN) and Transformer, to hierarchically capture interactive features over prior knowledge (agent trajectories and future lanes), and then obtain the structural representations of fine-grained future goals for per interacting agent, by aggregating interactive features from the individual, local and global levels.  
	\item For the goal interactive prediction and trajectory interactive prediction, we propose the Conditional Goal Prediction Network (CGPNet) and the Goal-oriented Trajectory Forecasting Network (GTFNet) by embedding the conditional prediction into the goal-oriented trajectory prediction method. 
	Based on CGPNet and GTFNet, the future diverse interactions between two interacting agents can be captured by the learned joint distribution. 
	\item In addition, a goal interactive loss is established for the CGPNet, which aims to better learn the joint probability distribution over future goal candidates for the two interacting agents.
	\item \textcolor{black}{
	Comparison experiments on Argoverse motion forecasting dataset, In-house cut-in dataset, and Waymo open motion dataset verify the superiority of our CGTP framework over the mainstream marginal prediction models and state-of-the-art conditional prediction model.
	} 
\end{itemize}

\section{Related Work}\label{sec:relatedWork}

In this section, we provide a detailed trajectory prediction literature review with a particular emphasis on deep learning methods from the perspective of context encoding, anchor-based prediction and conditional prediction, respectively.   

\subsection{Context Encoding}\label{sec:Contex Information Encoding}
There is a family of work on trajectory prediction via convolutional neural networks (CNNs) rendered input as a multi-channel rasterized bird's-eye view (BEV) image \cite{ MTP2019, TPCN2021}. \textcolor{black}{However, such rasterized approaches are difficult in modeling long-range interactions and representing continuous physical states. An popular alternative is to use a vectorized method.
With this approach, the history of agent motion is typically encoded via sequence modeling techniques like RNNs \cite{MATF2019}, while the elements of the road graphs are approximately treated as pairwise-linear segments with additional attributes such as current states and semantic type. Furthermore, the information aggregation techniques are utilized to learn relationships between the agent dynamics in the context of the road graph scenarios. Transformer \cite{Transformer2017} can be denoted as one popular choice for interaction-aware motion modeling based on the attention mechanism, capturing relationships via three different axes over timesteps, agents and road elements. For instance, \cite{zhao2020spatial} focuses on temporal encoding and decoding by applying a self-attention module for timesteps-axis, while \cite{Jean2019} employs a new multi-head attention architecture to complete interactions between all agents. Unlike past work using independent self-attention for each axis, SceneTransformer \cite{SceneTrans2021} is designed to handle the interactive modeling among timesteps, agents and road graph elements in a unified way. Alternatively, GNN-based methods have recently shown promise in motion prediction tasks, by learning interactive graph representations from vectorized features via operators like graph convolution and message passing \cite{GraphReview2018, Pedestrain2021, Chaochenzhuolei}.}
VectorNet \cite{VectorNet2020} introduces a hierarchical graph method that first processes agent histories and map features in the form of polylines and then fuses them using a global interactive graph. Different from VectorNet, LaneGCN \cite{LaneGCN2020} merely constructs lane graph using graph convolution before capturing all possible agent-map interactions. However, the interaction-aware models above concentrate their focus more on interactive modeling between coarse-scale objects such as agent past trajectories or lanes \cite{zhang2022trajgen}, rather than the fine-grained elements of map topology such as the goals.


\subsection{Anchor-based Multimodal Trajectory prediction}\label{sec:anchor-based method}
The multimodal trajectory prediction models are largely realized by anchor-based methods. This class of methods choose different types of possible intents, including a diverse of future trajectories, goals, relevant lanes and regions, to represent the modes of agent trajectory distribution. A family of studies leverage future trajectories as good prior anchors, and then produce the final trajectory predictions by using a learning-based method. For example, MultiPath \cite{MultiPath2019} and CoverNet \cite{CoverNet2019} generate a candidate set of predefined future trajectories as hypothesis proposals, while PRIME \cite{PRIME2021} and TPNet \cite{TPNet2020} produce feasible future trajectories based on the model-based generator instead. Further, TNT \cite{TNT2020} generates goal-oriented trajectories to diversify the prediction modes, with the goal anchor candidates sampled from the High-Definition (HD) map. Besides, since the topological structure of lanes can be thought of a guidance for the motion of drivers, a vast majority of recent work leverage a set of instance-level lane entities as spatial proposals to generate multimodal plausible trajectories \cite{LaPred2021, GoalNet2020}. Unlike these anchor-based models above, \cite{mmTransformer2021} constructs a novel region-based training method in order to cover all the possible prediction modes in the determined scene with limited training samples. This approach divides the surrounding space into a small number of regions in the first place, and then refines the prediction outcomes to a specific region where ground truth locates. In conclusion, these anchor-based approaches commonly have two stages including the anchor selecting and trajectory regression, which are trained end-to-end with stronger interpretability. Unfortunately, the anchor-based methods are largely used in the marginal prediction process until now.


\subsection{Conditional Trajectory Prediction}
As illustrated in Section I, the marginal prediction approach can be hardly applied in the interactive driving scenarios, since such a model ignores the fact that the action made by interacting agent $A$ in the future may have a critical effect on the interacting agent $B$ and vice versa. Hence, a minimal number of studies have made explorations on modeling joint future trajectories based on the conditional prediction models \cite{CBP2021, SceneTrans2021, mfp2019, Precog2019, sun2022m2i, ProspectNet, ILVM2020}. These methods output future agent trajectories by conditioning on other interacting agents' explicit future trajectories or implicit latent variables. 
By comparison, we develop a CGPNet to complete goal interactive prediction in priority based on the conditional method, which takes as queries the potential future goals of agent $A$ and predicts the probability distribution over the future goal candidates of agent $B$ conditioned on per query. 
Followed by this, we consider interactions over the future trajectories timestep-by-timestep, designing GTFNet to predict interactive future behaviors in a rollout manner. 


\textcolor{black}{\section{Problem Formulation}}\label{sec:backgound} 
\textcolor{black}{\subsection{Variable Definition}}\label{sec:formulation} 
\begin{figure*}[!t]
	\centering
	\begin{center}
		\scriptsize
		\includegraphics*[width=7.1in]{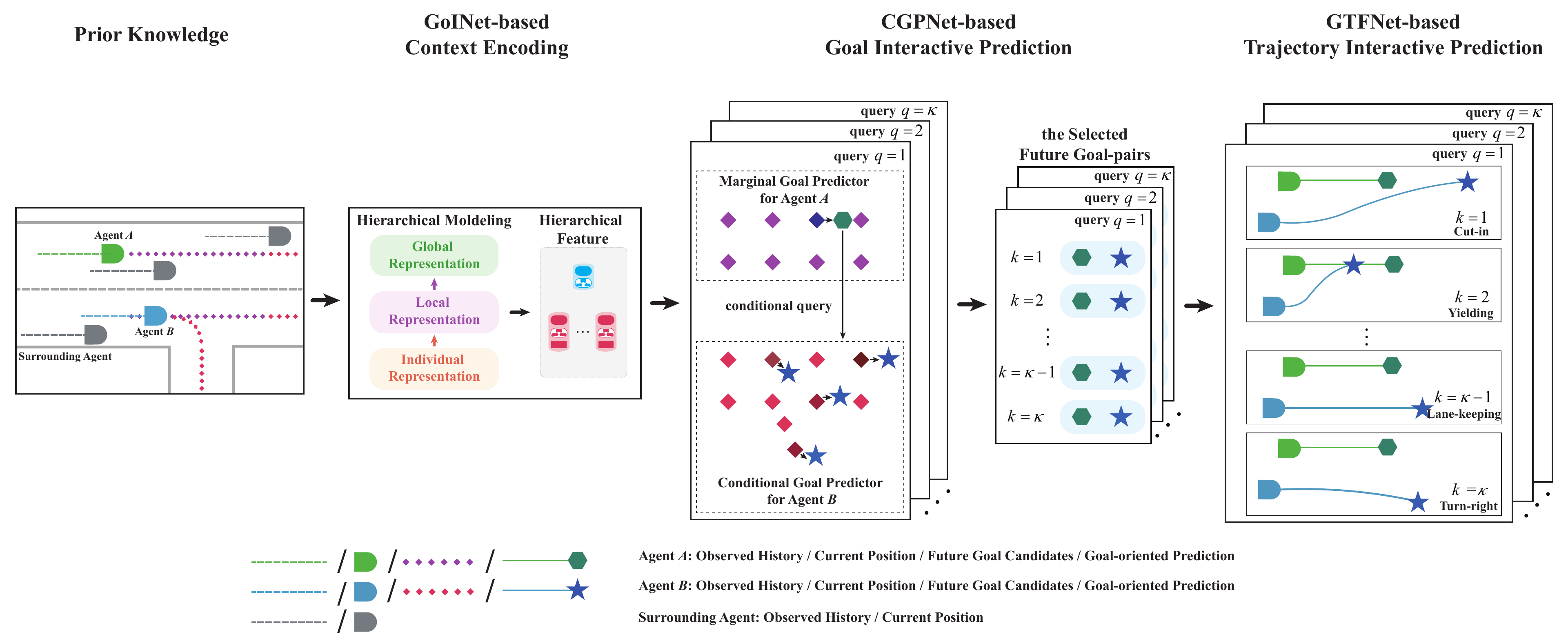}\\
		\caption{\textbf{An overview of the CGTP framework.} The proposed GoINet is first used to extract the hierarchical features over each interacting agent and its future goal candidates. Then, we select the interactive future goal-pairs via a novel CGPNet. Finally, the proposed GTFNet conducts the trajectory interactive prediction process to produce the goal-oriented predictions of two interacting agents, generating multimodal interactive behaviors such as cut-in, yielding, lane-keeping and turning right et al..}
		\label{fig:CGITP_framwork}
	\end{center}
	 \vspace{-1.3em} 	
\end{figure*}

\textcolor{black}{Given the scene information in a combined form as $\boldsymbol{C} = \boldsymbol{C}_A \cup \boldsymbol{C}_B$, our objective is to predict the future joint states $\boldsymbol{Y} = \boldsymbol{Y}_A \cup \boldsymbol{Y}_B$ of two interacting agents up to a finite horizon $T$, modeled as a joint distribution $p( \boldsymbol{Y} \mid \boldsymbol{C})$. 
\textcolor{black}{Towards each interacting agent $i$ \textcolor{black}{$\in$ $\{A, B\}$}, the scene information $\boldsymbol{C}_i: \{\boldsymbol{X}_i, \boldsymbol{L}_i\}$ contains dynamic and static representations \textcolor{black}{normalized at its reference frame}, where the agent trajectory set $\boldsymbol{X}_i=\{\boldsymbol{X}_i^{m}, m\in[0, O]\}$ includes the observed trajectory of predicted agent $\boldsymbol{X}_i^{0}$ and other agents' trajectories $\{\boldsymbol{{X}_i^{m}},{m\in[1,O]}\}$,
and $\boldsymbol{L}_i=\{\boldsymbol{L}_i^{m}, m\in[1, P]\}$ describes $P$ coarse-scale lanes that the agent $i$ is likely to reach in the future.}  }

\textcolor{black}{\subsection{Conditional Goal-oriented Trajectory Prediction}}\label{sec:formulation} 
For better comprehension, the marginal prediction method is first introduced, which lays a solid foundation for our proposed CGTP framework. In general, the marginal prediction methods are commonly built based on two assumptions.
\newtheorem{assumption}{Assumption}
\begin{assumption}\label{Agent Independence}
The agent's future states evolve independently from another interacting agent \cite{Desire2017, MultiPath2019, VectorNet2020, TNT2020}.
\end{assumption}

\textcolor{black}{Such independence assumption implies that marginal prediction methods predict the marginal distributions over individual agents without considering other interactions in the future.} \textcolor{black}{Once the \textit{Assumption \ref{Agent Independence}} is confirmed, the factorization over the joint distribution can be simplified as two marginal distributions}:
\begin{equation} \label{eq:n3-1}
\begin{gathered}
p(\boldsymbol{Y} \mid\boldsymbol{C}) = p( \boldsymbol{Y}_A \mid \boldsymbol{C}) p( \boldsymbol{Y}_B \mid \boldsymbol{C}).
\end{gathered}
\end{equation}
 Furthermore, \textcolor{black}{we adopt a goal-oriented trajectory prediction method TNT, a representative marginal prediction method, to effectively produce future trajectory with multimodality.}
 Towards each agent $i\in\{A,B\}$, the marginal distribution $p(\boldsymbol{Y}_i \mid \boldsymbol{C})$ can be decomposed based on future goal anchors, and then is marginalized over them:
\begin{equation} \label{eq:n3-2}
\begin{aligned}
p( \boldsymbol{Y}_i \mid \boldsymbol{C}) &= p(\boldsymbol{G}_{i}\mid \boldsymbol{C})p(\boldsymbol{Y}_i \mid \boldsymbol{G}_{i}, \boldsymbol{C}) \\
&= \sum_{\boldsymbol{g}_{i}^k \in \boldsymbol{G}_{i}}p(\boldsymbol{g}_{i}^k\mid \boldsymbol{C})p(\boldsymbol{Y}_i \mid \boldsymbol{g}_{i}^k, \boldsymbol{C}), 
\end{aligned}
\end{equation}
where $\boldsymbol{G}_i = \left\{\boldsymbol{g}_{i}^1, \boldsymbol{g}_{i}^2, \cdots, \boldsymbol{g}_{i}^K \right\}$ represents the location space of plausible future goal candidates for agent $i$, which captures $K$ uncertainties by relying on the known road information $\boldsymbol{L}_i$. 


\begin{assumption}\label{time Independence}
As for each agent $i$, the generation of future states is performed in an independent rollout manner \cite{TNT2020, Jean2019}.
\end{assumption}

Based on the \textit{Assumption \ref{time Independence}}, the future distribution for per agent $i$ can be factorized across time steps, by merely referring to its own previous states.
\begin{equation} \label{eq:n3-3}
\begin{gathered}
p( \boldsymbol{Y}_i \mid \boldsymbol{g}_{i}^k, \boldsymbol{C}) = \prod_{\delta=t+1}^{\delta=t+T} p(\boldsymbol{y}_i^{\delta} \mid \boldsymbol{y}_i^{t : \delta-1}, \boldsymbol{g}_{i}^k, \boldsymbol{C}),
\end{gathered}
\end{equation}
where $\boldsymbol{y}_i^\delta$ describes the future state of agent $i$ at time step $\delta$.

\textcolor{black}{Until now, the analysis above concludes that the goal-oriented trajectory prediction method ignores the future interaction during the joint trajectory prediction process. To bridge the gap between marginal prediction and interactive behavior prediction, we propose a novel CGTP framework by \textcolor{black}{considering conditional momdeling both in goal predictor and goal-oriented trajectory predictor}, 
collaboratively outputting scene-compliant joint future trajectories.
In this way, we first approximate the joint distribution as the factorization over a marginal distribution and a conditional distribution:}
\begin{equation} \label{eq:n4-10}
\begin{gathered}
p( \boldsymbol{Y} \mid \boldsymbol{C}) = p( \boldsymbol{Y}_A \mid \boldsymbol{C}) p( \boldsymbol{Y}_B \mid \boldsymbol{Y}_A, \boldsymbol{C}).
\end{gathered}
\end{equation}
\textcolor{black}{Different from Eq.~(\ref{eq:n3-1}), the factorization in Eq.~(\ref{eq:n4-10}) considers the agent $A$'s future intents have a potential influence on agent $B$. On one hand, the realization of agent $A$'s future trajectory can be roughly regarded as a marginal modeling process via a goal-oriented trajectory prediction method. On the other hand, we further make an approximate assumption to implement the conditional trajectory prediction for agent $B$. }

\begin{figure*}[htbp]
	\centering
	\begin{center}
		\scriptsize
		\includegraphics*[width=7.1in]{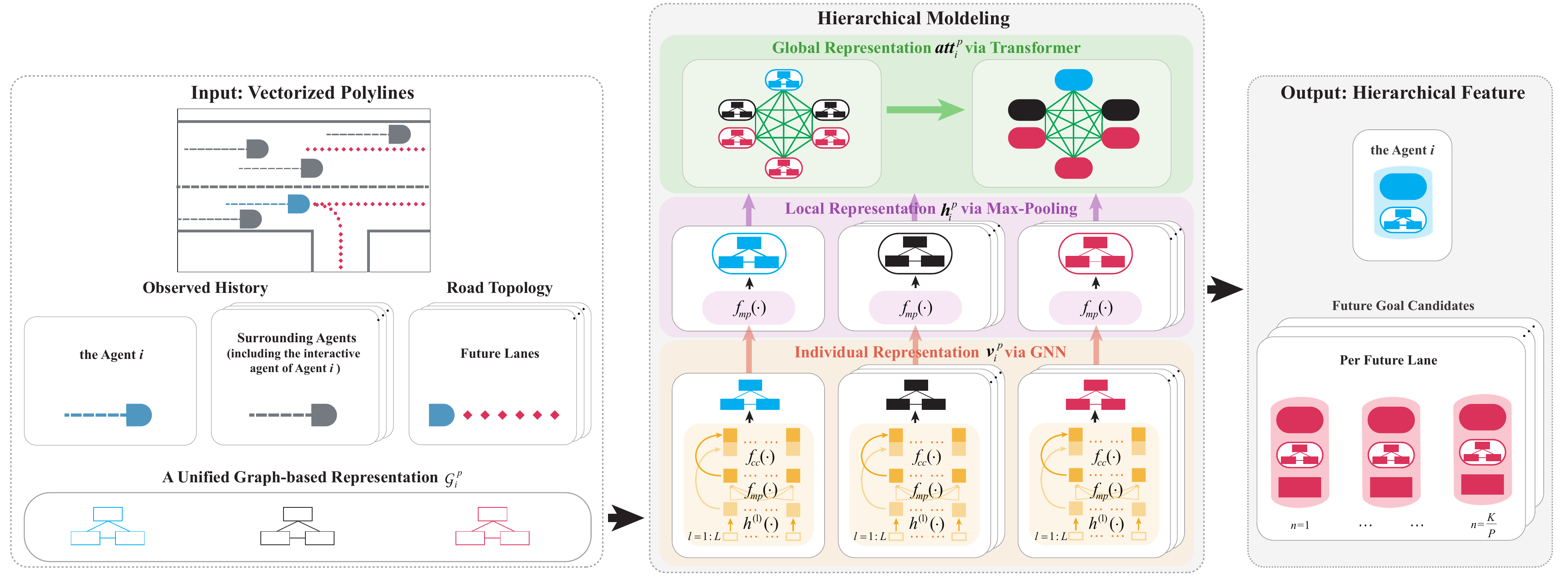}
		\caption{\textbf{The structure of the GoINet.} Towards each interacting agent, a unified graph-based representation is first formulated based on the scene information at its reference frame. Then, the hierarchical interactions are modeled from the individual, local and global three levels. Finally, we obtain the hierarchical features over the interacting agent and its fine-grained future goal candidates by means of a concatenation operator.}
		\label{fig:GoINet}
	\end{center}
   \vspace{-1.3em} 
\end{figure*}

\begin{assumption}\label{Conditional Dependence}
For interactive behavior prediction, the conditional distribution over agent $B$ can be largely determined by the agent $A$'s future goals instead of its overall future trajectories.
\end{assumption}

Based on the \textit{Assumption \ref{Conditional Dependence}} and the goal-oriented trajectory prediction method, the conditional distribution over agent $B$ is decomposed in a similar manner of agent $A$:
\begin{equation} \label{eq:n4-11}
\begin{gathered}
p(\boldsymbol{Y}_B \mid \boldsymbol{Y}_A, \boldsymbol{C}) = p( \boldsymbol{Y}_B \mid \boldsymbol{G}_A, \boldsymbol{C}) \\ = p(\boldsymbol{G}_{B} \mid \boldsymbol{G}_{A}, \boldsymbol{C})p(\boldsymbol{Y}_B \mid \boldsymbol{G}_{B}, \boldsymbol{G}_{A}, \boldsymbol{C}) \\
= \sum_{\small{\boldsymbol{g}_{A}^q, \boldsymbol{g}_{B}^k \in \boldsymbol{G}}}p(\boldsymbol{g}_{B}^k \mid \boldsymbol{g}_{A}^q, \boldsymbol{C})p(\boldsymbol{Y}_B \mid \boldsymbol{g}_{B}^k, \boldsymbol{g}_{A}^q, \boldsymbol{C}).
\end{gathered}
\end{equation}
\textcolor{black}{\textcolor{black}{There exits an obvious difference in the goal prediction process between Eq.~(\ref{eq:n3-2}) and Eq.~(\ref{eq:n4-11})}, indicating that the conditional modeling over agent $B$ aims to tackle the pairwise interactive trajectory prediction problem with the conditional modeling over future goal candidates, as described by $p(\boldsymbol{g}_{B}^k \mid \boldsymbol{g}_{A}^q, \boldsymbol{C})$.}
Here, in order to distinguish the indexes of future goal candidates for two interacting agents, we use $q$ to describe the index of the future goal candidates for agent $A$ instead. 

\textcolor{black}{Besides, our proposed CGTP framework conducts the goal-oriented trajectory prediction for each interacting agent via conditional modeling by referring to \cite{mfp2019}.} In the following, we take the interacting agent $A$ as an instance to describe the realization process of trajectory forecasting:
\begin{equation} \label{eq:n4-18}
\begin{aligned}
p( \boldsymbol{Y}_A \mid \boldsymbol{g}_{A}^q, \boldsymbol{C}) = \prod_{\delta=t+1}^{\delta=t+T} p(\boldsymbol{y}_A^{\delta} \mid \boldsymbol{y}^{t : \delta-1}, \boldsymbol{g}_{A}^q, \boldsymbol{C}) \\ =
\prod_{\delta=t+1}^{\delta=t+T} p(\boldsymbol{y}_A^{\delta} \mid \boldsymbol{y}_{A}^{t : \delta-1}, \boldsymbol{y}_{B}^{t : \delta-1}, \boldsymbol{g}_{A}^q, \boldsymbol{C}).
\end{aligned}
\end{equation}
As shown in Eq.~(\ref{eq:n4-18}), the conditional trajectory distribution of each interacting agent is explicitly dependent on its own future goal but implicitly dependent on the predicted future states of another interacting agent, which can be considered as a trajectory interactive prediction process in a step-wise rollout manner. 
\textcolor{black}{\section{Methodology}}\label{sec:ETCNet}

 An overview of our CGTP framework is shown in Fig.~\ref{fig:CGITP_framwork}. In the following, we first present the GoINet which summarizes the structural interactive representations over fine-grained future goal candidates.
 Then, we develop a CGPNet to conduct future interactions at the goal-based level, and to select the goal-pair candidates with future interactive intents. Further, a GTFNet is developed to generate goal-oriented trajectory-pairs in a step-wise rollout manner. Finally, we introduce the optimization process of our CGTP framework.

\subsection{GoINet-based Context Encoding}\label{sec:architecture}
The GoINet has three core steps: (1) establish a unified graph-based formulation for two typical types of vectorized representations, $i.e.$, agent history trajectories and future lanes; (2) leverage GNN, max-pooling and Transformer to construct hierarchical interactions at individual, local and global levels, respectively; (3) concatenate the features from three levels above to obtain the structural features over fine-grained future goal candidates, as shown in Fig.~\ref{fig:GoINet}.

 \textbf{Graph-based Representation Formulation.} \textcolor{black}{Inspired by VectorNet \cite{VectorNet2020}, we first abstract the scene elements $\{X_i, L_i\}_{|i\in\{A,B\}}$ (including agent history trajectories and future lanes) as polylines $\mathcal{P}_i$. All of these polylines can be approximated as sequences of vectors: for future lanes, we uniformly sample key points from the polylines at the same spatial distance to approximately represent the fine-grained goals, and sequentially connect the neighboring key points into vectors; for agent history trajectories, we can just sample key points with a fixed temporal interval, and connect them into vectors. Each vector can be denoted by
\begin{equation}
\begin{gathered}
\boldsymbol{v}_{i}^{\mathcal{P}} = [\boldsymbol{d}_{start}, \boldsymbol{d}_{end}, j],
\end{gathered}
\end{equation}
where $\boldsymbol{d}_{start}$ and $\boldsymbol{d}_{end}$ are coordinates of the start and end point of the vector; $j$ is the integer index of polyline $\mathcal{P}_i^j$. Then, for each polyline, we build a local graph as $\boldsymbol{\mathcal{G}}_{i}^{\mathcal{P}}=(\boldsymbol{\mathcal{V}}_{i}^{\mathcal{P}}, \boldsymbol{\mathcal{E}}_{i}^\mathcal{P})$, where $\boldsymbol{\mathcal{V}}_{i}^{\mathcal{P}}$ denotes a set of nodes with vector features $\boldsymbol{v}_{i}^{\mathcal{P}}$ and $\boldsymbol{\mathcal{E}}_{i}^\mathcal{P}$ is a set of edges encoding pairwise relations between nodes.
}

\textbf{Modeling Hierarchical Interactions.} 
To extract the temporal-spatial and semantic locality of nodes, we first deploy a general GNN approach to extract the individual features of vectors in each polyline. Towards each local graph $\boldsymbol{\mathcal{G}}_{i}^{\mathcal{P}}$, we formulate the learning scheme of every node representation $\boldsymbol{v}^{(l)}  = (\boldsymbol{v}_{i}^{\mathcal{P}})^{(l)} \in  \mathbb{R}^{2 l d_h }$ with max-pooling operator $f_{mp}(\cdot)$ and concatenation operator $f_{cc}(\cdot)$ in the $l$-th layer as
\begin{equation} \label{eq:n4-4}
\begin{gathered}
\boldsymbol{v}^{(l)} = f_{cc}\left( \left\{ h^{(l)}\left(\boldsymbol{v}^{(l-1)}\right),  f_{mp} \left( \left\{  h^{(l)}\left(\boldsymbol{v'}^{(l-1)}\right) \right\} \right) \right\} \right), \\
\forall l \in [1, L],
\end{gathered}
\end{equation}
where $\boldsymbol{v'}$ denotes the remaining nodes in $\boldsymbol{\mathcal{V}}_{i}^{\mathcal{P}}$ except for $\boldsymbol{v}$; $d_h$ represents the initial dimension of the hidden units at the first layer of GNN. In addition, $h^{(l)}(\cdot)$ denotes a mapping function at the $l$-th layer to iteratively encode the individual node embedding, which shares weights across all nodes. The mapping function $h^{(l)}(\cdot)$ is realized by a single fully connected layer with  Layer Normalization \cite{LayerNorm2016} and ReLU non-linearity. Specifically, we initialize $(\boldsymbol{v}_{i}^{\mathcal{P}})^{(0)} = \boldsymbol{v}_{i}^{\mathcal{P}}$. 
After $L$ layers of aggregation, we obtain the individual feature of nodes in each local graph. Second, the local-level representation over each entire polyline can be summarized by 
\begin{equation} \label{eq:n4-5}
\begin{gathered}
\boldsymbol{h}_{i}^{\mathcal{P}} = f_{mp} \left( \left\{ \left(\boldsymbol{v}_{i}^{\mathcal{P}}\right)^{(L)} \mid \forall \boldsymbol{v}_{i}^{\mathcal{P}} \in \boldsymbol{\mathcal{V}}_{i}^{\mathcal{P}} \right\}\right),
\end{gathered}
\end{equation}
which models interactions among all nodes' individual representations in each local graph via max-pooling operator. More formally, we stack these local features into a matrix as $\boldsymbol{H}_i \in  \mathbb{R}^{(O+P) \times d_H}$, where $d_H = 2Ld_h$. 
\textcolor{black}{Finally, a Transformer layer is employed to draw global dependencies between the local-level features over agent trajectories and future lanes.} The output of self-attention computation for per interacting agent $i$ is described as follows 
\begin{equation} \label{eq:n4-6}
\begin{gathered}
\boldsymbol{Att}_i = {\rm{softmax}} \left( \frac{\boldsymbol{Q}_i \left(\boldsymbol{K}_i\right) ^ \mathrm{T}}{\sqrt{d_k}} \right) \boldsymbol{V}_i, 
\end{gathered}
\end{equation}
where each row of the matrix $\boldsymbol{Att}_i$ is a global feature of a specific polyline, i.e. $\boldsymbol{att}_{i}^{\mathcal{P}}$, and $d_k=d_H$.
\textcolor{black}{In Eq.~(\ref{eq:n4-6}), the set of queries $\boldsymbol{Q}_i$, keys $\boldsymbol{K}_i$ and values $\boldsymbol{V}_i$ are obtained by making linear projections to the local representation matrix $\boldsymbol{H}_i$.}


\textbf{Obtaining Structural Representations.} \textcolor{black}{As shown in the right part of Fig.~\ref{fig:CGITP_framwork}, the hierarchical encoding information $\boldsymbol{s}_i^X$ for per interacting agent $i$ is a concatenation of the local and global representations over its observed history $\boldsymbol{X}_i^0$. Since it is necessary to consider the nodes feature of future lanes in the representations of future goals, we combine future lane features from individual, local and global views to encode the fine-grained structural representations $\{\boldsymbol{s}_{i}^{g,k}, k \in [1, K]\}$ of the goal candidates $\{\boldsymbol{g}_{i}^{k}, k \in [1, K]\}$.}
In the end, we take as input the structural interactive features above to the following modules.

\subsection{CGPNet-based Goal Interactive Prediction}\label{sec:definition}

In this section, we focus on introducing an implementation paradigm for estimating the conditional probability distribution over the future goal candidates for agent $B$, $i.e.$ $p\left( \boldsymbol{g}_{B}^{k} \mid \boldsymbol{g}_{A}^{q}, \boldsymbol{C} \right)$, which takes as conditional queries the potential future goals from agent $A$. As illustrated in TNT \cite{TNT2020}, the determination of future goals relies on discrete future goal candidates $\boldsymbol{g}_{B}^{k}$ and their corresponding continuous offsets $\boldsymbol{\Delta} \boldsymbol{g}_{B}^{k}$ to the real endpoint $\boldsymbol{y}_{B}^{t+T}$. Hence, the conditional probability distribution over future goal candidates can be factorized into these two influential elements above: 
 
\begin{equation} \label{eq:n4-12}
\begin{aligned}
p\left( \boldsymbol{g}_{B}^{k} \mid \boldsymbol{g}_{A}^{q}, \boldsymbol{C} \right) &= \pi \left( \boldsymbol{g}_{B}^{k} \mid \boldsymbol{g}_{A}^{q}, \boldsymbol{C} \right) \\ &\cdot \mathcal{N}\left( \boldsymbol{\Delta} \boldsymbol{g}_{B}^{k} \mid \boldsymbol{\mu} \left( \boldsymbol{\Delta} \boldsymbol{g}_{B}^{k}\right), \boldsymbol{\Sigma} \left( \boldsymbol{\Delta} \boldsymbol{g}_{B}^{k}\right) \right),
\end{aligned}
\end{equation}
where $\pi \left( \boldsymbol{g}_{B}^{k} \mid \boldsymbol{g}_{A}^{q}, \boldsymbol{C} \right)$ describes the uncertainty across a candidate set of agent $B$'s future goals using a softmax distribution:
\begin{equation} \label{eq:n4-13}
\begin{gathered}
\pi \left( \boldsymbol{g}_{B}^{k} \mid \boldsymbol{g}_{A}^{q}, \boldsymbol{C} \right) = \frac{ {\rm exp} \ f_B^{seg}\left( \boldsymbol{s}_{B}^{g,k}, \boldsymbol{g}_{B}^{k}, \phi_{B} \left(\boldsymbol{g}_{A}^{q}\right) \right)}{\sum \limits_{ \small{\boldsymbol{g}_{B}^{ k'} }} {\rm exp} \ f_B^{seg}\left( \boldsymbol{s}_{B}^{g,k'}, \boldsymbol{g}_{B}^{k'}, \phi_{B} \left(\boldsymbol{g}_{A}^{q} \right)\right)}.
\end{gathered}
\end{equation} 
This conditional probability distribution is learned by a segmentation task.
Subsequently, we obtain the corresponding offset from a generalized Gaussian distribution $\mathcal{N}\left( \boldsymbol{\Delta} \boldsymbol{g}_{B}^k \mid \boldsymbol{\mu} \left( \boldsymbol{\Delta} \boldsymbol{g}_{B}^k\right), \boldsymbol{\Sigma}\left( \boldsymbol{\Delta} \boldsymbol{g}_{B}^k\right)\right)$, where $\boldsymbol{\mu} \left( \boldsymbol{\Delta} \boldsymbol{g}_{B}^k\right)$ denotes mean as 
\begin{equation} \label{eq:n4-14}
\begin{gathered}
\boldsymbol{\mu} \left( \boldsymbol{\Delta} \boldsymbol{g}_{B}^{k}\right) = f_B^{reg}\left( \boldsymbol{s}_{B}^{g,k}, \boldsymbol{g}_{B}^k, \phi_B \left(\boldsymbol{g}_{A}^q\right) \right)
\end{gathered}
\end{equation} 
which is primarily modeled by a regression task. Besides, let $\boldsymbol{\Sigma} \left( \boldsymbol{\Delta} \boldsymbol{g}_{B}^{k}\right)$ denote variance assumed to be an identity matrix in this paper. From another view, both $f_B^{seg}(\cdot)$ and $f_B^{reg}(\cdot)$ are implemented by a three-layer multilayer perceptron (MLP) to predict the conditional distribution and offsets over future goal candidates. Concretely, the input of these two mapping functions is mainly derived from two aspects. One class of input is related to each future goal candidate $\boldsymbol{g}_{B}^k$ and its corresponding structural representation $\boldsymbol{s}_{B}^{g,k}$ extracted from the GoINet. Furthermore, an agent $B$-centric transformation function $\phi_{B} (\cdot)$ is deployed to acquire the potential future goals $\boldsymbol{g}_{A}^q$ of agent $A$ at the agent $B$'s reference frame, which we refer to as conditional queries, as well as another class of input. 

\textcolor{black}{Before obtaining conditional probability distribution upon Eq.~(\ref{eq:n4-12}), we also require estimating the marginal probability distribution over the future goal candidates for agent $A$, named $p\left( \boldsymbol{g}_{A}^{q} \mid \boldsymbol{C}\right)$, by turning off inputs from the conditional query in the model. Thus, the marginal probability distribution is described by the simplified expressions $\pi \left( \boldsymbol{g}_{A}^{q} \mid \emptyset, \boldsymbol{C} \right)$ and $\boldsymbol{\mu} \left( \boldsymbol{\Delta} \boldsymbol{g}_{A}^{q} \mid \emptyset \right)$, and the realizations over them can approximately refer to Eq.~(\ref{eq:n4-13}) and Eq.~(\ref{eq:n4-14}).}

Given the marginal and conditional probability distribution for the goal interactive prediction process, we now can compute the joint probability distribution $p\left(\boldsymbol{g}_{B}^{k}, \boldsymbol{g}_{A}^{q} \mid \boldsymbol{C}\right)$. For simplification, suppose that the joint probability distribution above can be approximately replaced by $\pi \left( \boldsymbol{g}_{B}^{k}, \boldsymbol{q}_{A}^{q} \mid \boldsymbol{C}\right)$, described as 
\begin{equation} \label{eq:n4-17}
\begin{gathered}
p\left( \boldsymbol{g}_{B}^{k}, \boldsymbol{g}_{A}^{q} \mid \boldsymbol{C}\right) = \pi\left( \boldsymbol{g}_{B}^{k}, \boldsymbol{g}_{A}^{q} \mid \boldsymbol{C}\right) \\
= \pi\left( \boldsymbol{g}_{B}^{k} \mid \boldsymbol{g}_{A}^{q}, \boldsymbol{C}\right) \pi\left( \boldsymbol{g}_{A}^{q} \mid \emptyset, \boldsymbol{C}\right).
\end{gathered}
\end{equation}

\subsection{GTFNet-based Trajectory Interactive Prediction}\label{sec:definition}
After obtaining a candidate set of future goal-pairs, we then build a trajectory interactive prediction module to output joint future trajectories in a synchronized rollout manner.
Different from Eq.(\ref{eq:n3-3}), this module predicts the joint states of two interacting agents at time step $\delta + 1$ by taking into account each other's predicted state at time step $\delta$. 
Given the determined future goal, we take agent $A$ as an instance to introduce the implementation paradigm over the unimodal conditional distribution for future trajectory, as described by Eq.(\ref{eq:n4-18}). We design a GRU-based encoder-decoder neural network to realize the generation of future trajectory, which shares the trainable parameters across two interacting agents. In detail, both encoder and decoder use GRU mapping $f_{gru}(\cdot)$ to recursively update the hidden unit along the temporal axis. Especially, the input representations of GRU mapping are defined by different forms in terms of encoding and decoding necessity. On one hand, the encoding GRU captures temporal relationships by taking as input the observed history $\boldsymbol{X}_i^0$. On the other hand, the decoding GRU updates the hidden state and then a trajectory predictor $f_{traj}(\cdot)$, implemented by 1-layer MLP, is followed to predict the future location at the current time step, which is transformed via $\phi_{B}(\cdot)$ and subsequently referred as input to the prediction process of agent $B$ at the next future time step. In the meanwhile, the concatenation of a goal candidate $\boldsymbol{g}_{A}^{q}$ and hierarchical interactive representation $\boldsymbol{s}_A^X$ is also served as input to determine the future intent at which agent $A$ will arrive. 


\subsection{Optimization Design for CGTP Framework}\label{sec:algorithm}
The proposed CGTP framework is trained via supervised learning in an end-to-end way. 
The total learning loss function contains goal prediction loss and trajectory prediction loss, defined as:
\begin{equation} \label{eq:n4-19}
\begin{gathered}
L^{total} = L^{g} + L^{traj}.
\end{gathered}
\end{equation}
In the following, we illustrate the training strategy of the two components above from interacting agents and joint modes in two aspects. Besides, a detailed training algorithm pseudocode is provided as 
described in Algorithm~\ref{alg:algrithm1}.

\textbf{Training on Goal Interactive Prediction.} To effectively model the joint distribution of future goal candidates, our goal prediction loss $L^{g}$ is modified to supervise the general intersection of future goal candidate sets from the two interacting agents, in addition to both of the single interacting agent's future goal candidate set. Thus, the goal prediction loss can be decomposed into three parts in sequential order according to their single forms and joint form:      
\begin{equation} \label{eq:n4-20}
\begin{gathered}
L^{g} =L_A^{g} + L_B^{g} + L_{Joint}^{g}.
\end{gathered}
\end{equation}

First, We introduce the marginal goal prediction loss $L_A^g$ of agent $A$. On one hand, the binary cross entropy loss $L_{BCE}(\cdot)$ is used to learn the marginal probability distribution $\pi \left( \boldsymbol{g}_{A}^{q} \mid \emptyset, \boldsymbol{C}\right)$.
On the other hand, the mean square error loss $L_{MSE}(\cdot)$ is employed to learn the offset mean $ \boldsymbol{\mu} \left( \boldsymbol{\Delta} \boldsymbol{g}_{A}^{q} \mid \emptyset \right)$ instead.
Hence, the marginal goal prediction loss $L_A^g$ is represented by 
\begin{equation} \label{eq:n4-21}
\begin{gathered}
L_A^{g} = \frac{1}{K}\sum_{q=1}^{K} \left[ L_{BCE}\left( \pi\left(\boldsymbol{g}_{A}^{q} \mid \emptyset, \boldsymbol{C}\right), \mathbb{1}\left(q \in \boldsymbol{K}_A\right)\right) 
\right.
\\
\left.
+ L_{MSE}\left( \boldsymbol{\mu}\left(\boldsymbol{ \Delta} \boldsymbol{g}_{A}^{q} \mid \emptyset \right), \mathbb{1}\left(q \in \boldsymbol{K}_A\right) \boldsymbol{\Delta} \boldsymbol{g}_{A}^{q}\right)\right],
\end{gathered}
\end{equation}
where $\mathbb{1}(\cdot)$ is an indicator function, and $\boldsymbol{K}_A$ represents the index set of goal candidates which cover the top $\mathcal{K}$ candidates closest to the real endpoint $\boldsymbol{y}_A^{t+T}$ of agent $A$. 

Subsequently, once determining the top $\mathcal{K}$  agent $A$'s future modes as estimated by $\pi \left( \boldsymbol{g}_{A}^{q} \mid \emptyset, \boldsymbol{C}\right)$, we accordingly optimize the conditional goal prediction loss over them for another interacting agent $B$:
 \begin{equation} \label{eq:n4-22}
 \begin{gathered}
 L_B^{g} = 
 \frac{1}{\mathcal{K} \cdot K}\sum_{q=1}^{\mathcal{K}}\sum_{k=1}^{K} \left[ L_{BCE}\left( \pi\left(\boldsymbol{g}_{B}^{k} \mid \boldsymbol{g}_{A}^{q}, \boldsymbol{C} \right), \mathbb{1}\left(k \in \boldsymbol{K}_B\right)\right) 
 \right.
 \\
 \left.
 + L_{MSE}\left( \boldsymbol{\mu}\left(\boldsymbol{ \Delta} \boldsymbol{g}_{B}^{k}\right), \mathbb{1}\left(k \in \boldsymbol{K}_B\right) \boldsymbol{\Delta} \boldsymbol{g}_{B}^{k}\right)\right].
 \end{gathered}
 \end{equation}
\noindent Furthermore, to enable a smooth training process, we employ a teacher forcing technique \cite{teacherforcing1989} by using the real endpoint of agent $A$ as the conditional query. Similarly, we correspondingly obtain the top $\mathcal{K}$ potential goal candidates of the agent $B$ at each conditional mode $q$, and then $\mathcal{K}^2$ goal-pairs begin to emerge that reflect different future interactive intents. 

Finally, we design a novel goal interactive loss to accurately learn the joint probability distribution among the two classes of selected goal candidate sets:
 \begin{equation} \label{eq:n4-23}
 \begin{gathered}
 L_{Joint}^{g} = \frac{1}{\mathcal{K}^2}\sum_{q=1}^{\mathcal{K}}\sum_{k=1}^{\mathcal{K}}  L_{BCE}\left( \pi\left(\boldsymbol{g}_{B}^{k}, \boldsymbol{g}_{A}^{q} \mid \boldsymbol{C}\right), \mathbb{1}\left(\kappa = k^{J}\right)\right), \\
 \kappa = \mathcal{K}(q-1)+k,
 \end{gathered}
 \end{equation}
\noindent where $\kappa$ represents the index of the selected goal-pairs, which also denotes the index of joint modes for the goal-oriented trajectory-pairs later. Different from Eq.(\ref{eq:n4-21}) and Eq.(\ref{eq:n4-22}), $k^{J}$ is the index of a specific case where both two agents' future goal candidates most closely match their corresponding ground truth endpoints.

\textbf{Training on Trajectory Interactive Prediction.} After attaining $\mathcal{K}^2$ diverse combinations of future intents, we accordingly obtain their $\mathcal{K}^2$ goal-oriented trajectory-pairs. Let $\hat{\boldsymbol{Y}}_{i}^{\kappa} =  \left\{ \hat{\boldsymbol{y}}_{i}^{\kappa,t+1},  \hat{\boldsymbol{y}}_{i}^{\kappa,t+2}, \cdots, \hat{\boldsymbol{y}}_{i}^{\kappa,t+T} \right\}$ represent the goal-oriented trajectory prediction of interacting agent $i$ at joint mode $\kappa$. We also adopt the mean square error loss to minimize the Euclidean distance between the most likely predicted joint states and the ground truth  at per future time step:

 \begin{equation}
 \label{eq:n4-24}
 \begin{gathered}
 L^{traj} = \frac{1}{2\mathcal{K}^2 \cdot T}\sum_{\kappa=1}^{\mathcal{K}^2}\sum_{\delta=t+1}^{t+T} \sum_{i} L_{MSE}\left( \hat{\boldsymbol{y}}_{i}^{\kappa,\delta}, \mathbb{1}\left(\kappa = k^{J}\right) \boldsymbol{y}_i^{\delta}\right).
 \end{gathered}
 \end{equation}
 \noindent 
Moreover, the teacher forcing approach is utilized during the trajectory predictive rollouts as well by feeding one agent's ground truth observation at time step $\delta$, served as a conditional interaction, to another agent at time step $\delta+1$. In addition, during the training time, we also consider the real endpoint of each interacting agent as the goal anchor to guide the prediction of the goal-oriented future trajectory.
  
 At inference process, we replace the real endpoint of agent $A$ with its predicted future goals, served as conditional queries, to estimate conditional distribution over the future goal candidates for agent $B$. In addition, during the trajectory interactive prediction process, each interacting agent generates trajectories by considering its predicted future goals as anchors instead of its real endpoint. Also, the future ground truth is substituted by the predicted states to instruct the future interactive rollouts between two interacting agents in a step-wise manner.

\begin{algorithm*}[t]
	\caption{{Optimization for CGTP Framework}}
	\label{alg:algrithm1}
	\LinesNumbered

	\textbf{{Input:}} 
	scene information $\boldsymbol{C}=\left(\boldsymbol{C}_A, \boldsymbol{C}_B \right)$ and future joint states $\boldsymbol{Y}=\left(\boldsymbol{Y}_A, \boldsymbol{Y}_B \right)$
	
	\textbf{{Initialize:}} Randomly initialize GoINet's parameter $\theta_{GoINet}$, the CGPNet's all parameters $\theta_{A}^{seg}$, $\theta_{A}^{reg}$, $\theta_{B}^{seg}$, $\theta_{B}^{reg}$ and the GTFNet's all parameters $\theta^{Enc}_{gru}$, $\theta^{Dec}_{gru}$, and $\theta_{traj}$
	
	\While{  \rm{not \ convergence}}
	{
		\For{agent $i$ in \{A, B\}}
		{$\boldsymbol{s}_i^X \gets$ encode the structural representation over history states via GoINet
			
		$\boldsymbol{s}_i^{g, n} \gets$ encode the structural representations over future goal candidates via GoINet
		
		 $\boldsymbol{u}_i^X \gets f_{gru}^{Enc}(\boldsymbol{X}_i^0 \mid \theta^{Enc}_{gru}) $: encode the temporal representation over history states via GRU-based encoder}

		\textbf{Training on the marginal prediction over future goal candidates for agent $A$}
		
		$ \pi \left( \boldsymbol{g}_{A}^{q} \mid \emptyset, \boldsymbol{C}; \theta_A^{seg}\right) \gets$ estimate the marginal probability distribution 
		
		$\boldsymbol{\mu}\left( \boldsymbol{\Delta} \boldsymbol{g}_{A}^{q} \mid \emptyset;  \theta_A^{reg}\right) \gets$ predict the offset mean 
		
		Update $\theta_{GoINet}$, $\theta_{A}^{seg}$ and  $\theta_{A}^{reg}$ by minimizing $L_A^g$ 
		
		\For{ condional queries $q = 1 \ to \ \mathcal{K} $ }
		{
			\textbf{Training on the conditional prediction over future goal candidates for agent $B$ at per query $q$}
			
			
			$ \pi \left( \boldsymbol{g}_{B}^{k} \mid \boldsymbol{g}_{A}^{q}, \boldsymbol{C}; \theta_B^{seg}\right) \gets$ estimate the conditional probability distribution 
			
			$\boldsymbol{\mu}\left( \boldsymbol{\Delta} \boldsymbol{g}_{B}^{k} \mid  \theta_B^{reg}\right) \gets$ predict the offset mean 
			 
			 \textbf{Training on the joint prediction over $\mathcal{K}$ goal-pair candidates at per query $q$}
			 
			 $\pi \left( \boldsymbol{g}_{B}^{k}, \boldsymbol{g}_{A}^{q} \mid \boldsymbol{C}; \theta_A^{seg}, \theta_B^{seg}\right) \gets$ estimate the joint probability distribution 

			\For{ conditional intents $k= 1 \ to \ \mathcal{K} $ }
			{
				$\kappa = \mathcal{K}(q-1) + k \gets$ calculate the prediction mode
				
				\textbf{Training on the joint prediction over goal-oriented future trajectories at per mode $\kappa$ }
				
				\For{ $Timesteps \ \delta= t+1 \ to \ t+T $ }
				{
					
					$\boldsymbol{r}_{A}^{\kappa,\delta} \gets f^{Dec}_{gru}\left( \phi_A \left( \hat{\boldsymbol{y}}_{B}^{\kappa,\delta-1}\right), \boldsymbol{g}_{A}^{q}, \boldsymbol{s}_A^X, \boldsymbol{u}_A^X \mid \theta^{Dec}_{gru}\right)$
					
					
					$\boldsymbol{r}_{B}^{\kappa,\delta} \gets f^{Dec}_{gru}\left( \phi_B \left( \hat{\boldsymbol{y}}_{A}^{\kappa,\delta-1}\right), \boldsymbol{g}_{B}^{k}, \boldsymbol{s}_B^X, \boldsymbol{u}_B^X \mid \theta^{Dec}_{gru}\right)$
					
					$\hat{\boldsymbol{y}}_{A}^{\kappa,\delta}, \hat{\boldsymbol{y}}_{B}^{\kappa,\delta} \gets f_{traj}\left(\boldsymbol{r}_{A}^{\kappa,\delta}, \boldsymbol{r}_{B}^{\kappa,\delta} \mid \theta_{traj} \right)$
					}
				}
		}
		Update $\theta_{GoINet}$, $\theta_B^{seg}$ and $\theta_B^{reg}$ by minimizing $L_B^g$ 
		
		Update $\theta_{GoINet}$, $\theta_A^{seg}$ and $\theta_B^{seg}$ by minimizing $L_{Joint}^g$ 
		
		Update $\theta_{GoINet}$, $\theta^{Enc}_{gru}$, $\theta^{Dec}_{gru}$ and $\theta_{traj}$ by minimizing $L^{traj}$ 
	}
	The $\mathcal{K}^2$ multimodal futue trajectories $\left\{ \hat{\boldsymbol{Y}}_{i}^1, \hat{\boldsymbol{Y}}_{i}^2, \cdots, \hat{\boldsymbol{Y}}_{i}^{\mathcal{K}^2}\right\}$ are obtained for per interacting agent $i$
	
	\textbf{return} All trainable parameters of GoINet, CGPNet and GTFNet
\end{algorithm*}

\section {Experiment}\label{sec:Experiment}
In this section, we first introduce the experimental settings, including datasets, metrics and implementation details. Subsequently, we compare our CGTP framework against the existing trajectory prediction methods.
In addition, the ablation studies are conducted to validate the effectiveness of the key design for our novel approach.  In the end, some qualitative analysis are performed in the multimodality and future interactivity aspects.

\subsection{Experimental Settings}
\emph{(1) Datasets:} \textcolor{black}{We evaluate our CGTP framework on three large-scale complex driving datasets: Argoverse motion forecasting dataset \cite{Argoverse2019}, In-house cut-in dataset and Waymo open motion dataset \cite{WaymoDataset2021}.} 

\noindent\textbf{Argoverse motion forecasting dataset} is a widely-used trajectory prediction dataset recorded in over 30K traffic scenarios from Pittsburgh and Miami. These scenarios produce a series of frames sampled at 10Hz, which are further split into training and validation sets with 205942 and 39472 frames, respectively. Different from prior literature, our work focuses on joint trajectory prediction for two agents. Thus, given the positions of all agents in each frame within the past 2 seconds observation, we consider the two interesting agents, with type 'agent' and 'av', as agent $A$ and agent $B$, separately, whose 3 seconds future trajectories need to be evaluated. Besides, this dataset provides a friendly interface to conveniently retrieve lane segments and their connection relationships for each frame. However, one limitation of this dataset is that there exist rare scenarios where the two agents interact with each other in the future. \textcolor{black}{To overcome this issue, the interactive datasets are taken into account as follows.}

\noindent\textbf{In-house cut-in dataset} is a Baidu internal dataset to support the trajectory interactive prediction task with two interacting agents. This large-scale dataset was collected in the specific cut-in scenarios from Beijing, China, which is divided into two branches in terms of junction and non-junction environments. On one hand, towards the junction environment, there are 180201 interactive frames in total extracted from over 11K unique cut-in scenarios, which are then split into 162381 frames for training and 17820 frames for validation. On the other hand, we provide 193401 interactive frames recorded in more than 12K non-junctional cut-in scenarios, while the training and validation sets contain 162556 and 30845 frames, saparately. 
Further, in this paper, the interactive pair is achieved by agent $A$ and $B$, and we choose the agent $A$ as the query agent to influence the cut-in reactions for agent $B$. Given 2 seconds of observed history, our objective is to predict 3 seconds of joint future trajectories for two interacting agents in each cut-in frame. The agent trajectories are sampled at 10Hz and the road topology are provided in the form of centerlines and lane boundaries.

\noindent\textcolor{black}{\textbf{Waymo open motion dataset} (WOMD) is by far the most diverse interactive motion dataset to the best of our knowledge. It contains more than 570 hours of unique data over 1750 km of roadways. Since WOMD provides specific labels for interacting vehicles, 158810 and 33686 interactive frames can be extracted from the training and validation dataset, respectively. Further, we leverage the relation predictor in M2I \cite{sun2022m2i} to provide the influencer-reactor relationships between each interactive pair, with the influencer and reactor determined as agent $A$ and $B$, separately. Given 1.1 seconds of agent states sampled at 10 Hz, we focus on the interactive prediction task to predict the joint future positions of two interacting agents for the next 8 seconds in the future. In addition to history trajectories, the map features, represented by lane polylines, are also included in the prior observations of each frame.}

\emph{(2) Metrics:} By referring to the evaluation setting from \cite{WaymoDataset2021} and \cite{Argoverse2019}, we use minimum Average Displacement Error (minADE), minimum Final Displacement Error (minFDE) and Miss Rate (MR) to measure distance error over joint trajectory predictions, and these metrics are applied to the three datasets. Towards WOMD and the In-house dataset, we also report the Overlap Rate (OR) to measure the collision frequency between two interacting agents. \textcolor{black}{Besides, mean Average Precision (mAP) is considered in WOMD to measure the quality of confidence score over joint predictions, which is the official ranking metric used by WOMD benchmark.} On the other hand, due to the unique cut-in attributes recorded from the In-house dataset, we design the Cut-in Rate (CR) metric to identify whether the agent $B$ has the ability to early merge to that lane where agent $A$ keeps, on the condition that no collision occurs. In the following, we give detailed expressions of these metrics above.

\noindent\textbf{minADE}. The minimum Average Displacement Error is defined as the $\ell_2$ distance between the ground truth and the closest predicted trajectory-pair:
 \begin{equation} \label{eq:n5-1}
 \begin{gathered}
minADE=\frac{1}{2T} \min\limits_{\kappa}\sum_{i}\sum_{\delta=t+1}^{t+T}  \Vert \hat{\boldsymbol{y}}_{i}^{\kappa,\delta}-\boldsymbol{y}_i^{\delta}  \Vert _ 2.
 \end{gathered}
 \end{equation}
 
\noindent\textbf{minFDE}.  The minimum Final Displacement Error is obtained by merely computing the minADE at the last future time step:
 \begin{equation} \label{eq:n5-2}
 \begin{gathered}
 minFDE=\frac{1}{2} \min\limits_{\kappa}\sum_{i}  \Vert \hat{\boldsymbol{y}}_{i}^{\kappa, t+T}-\boldsymbol{y}_i^{t+T}  \Vert _ 2.
 \end{gathered}
 \end{equation}
 
 \noindent\textbf{MR}. The Miss Rate is created by calculating an indicator function $IsMiss(\cdot)$ for each frame in turn and then averaging over all dataset. \textcolor{black}{For a specific frame, a miss is assigned if none of the joint predictions are within the given threshold(s) of the ground truth:}       
  \textcolor{black}{\begin{equation} \label{eq:n5-3}
  \begin{gathered}
  IsMiss(\cdot) =  {\rm{min}}_{\kappa} \vee_{i} \mathbb{1}\left(Dist_{i}^{\kappa} > Dist_{thre} \right),
  \end{gathered}
  \end{equation}}
\textcolor{black}{where $Dist_{i}^{\kappa}$ and $Dist_{thre}$ have different definitions in terms of different datasets. For Argoverse and In-house datasets, $Dist_i^{\kappa}$ calculates the final displacement error between ground truth and the future trajectory of agent $i$ at the joint mode $\kappa$, and $Dist_{thre}$ is a single distance threshold that is set to 2. Different from above, WOMD adopts different criteria for lateral deviation versus longitudinal depending on the initial velocity of the predicted agents. In this way, $Dist_{i}^{\kappa}$ and $Dist_{thre}$ are set as trajectory displacement errors and thresholds in the lateral and longitudinal two aspects.} 

  \begin{table*}[htbp]
  \centering
  \caption{Comparison with marginal prediction methods and ablations on Argoverse motion forecasting dataset}
    \begin{tabular}{c|c|ccc}
    \toprule
    \textbf{Joint Prediction} & \textbf{Methods} & \textbf{minADE $\downarrow$} & \textbf{minFDE $\downarrow$} & \textbf{MR $\downarrow$} \\
    \midrule
    \multicolumn{1}{c|}{\multirow{4}[2]{*}{\makecell[c]{Marginal \\ Predictions}}} & LSTM-ED \cite{Argoverse2019}   & 1.2221  & 2.7970  & 0.6868  \\
          & VectorNet (noGNN) \cite{VectorNet2020}  & 1.1327  & 2.6005  & 0.6522  \\
          & VectorNet \cite{VectorNet2020}  & 1.0959  & 2.3197  & 0.5592  \\
          & TNT \cite{TNT2020}    & 1.1320  & 2.5341  & 0.5664  \\
    \midrule
    \multicolumn{1}{c|}{\multirow{2}[2]{*}{\makecell[c]{Ablations}}} & CGTP (wo interactive loss) & 1.0063  & 2.2847  & 0.4900  \\
          & CGTP (w interactive loss) & \textbf{0.7533} & \textbf{1.6140} & \textbf{0.3369} \\
    \bottomrule
    \end{tabular}%
  \label{tab:Argoverse}%
\end{table*}%

\begin{table*}[htbp]
  \centering
  \caption{Comparison with marginal prediction methods and ablations on In-house cut-in dataset}
    \begin{tabular}{c|c|c|ccccc}
    \toprule
    \textbf{Scenarios} & \textbf{Joint Prediction} & \textbf{Methods} & \textbf{minADE $\downarrow$} & \textbf{minFDE $\downarrow$} & \textbf{MR $\downarrow$} & \textbf{OR $\downarrow$} & \textbf{CR $\uparrow$}\\
    \midrule
    \multicolumn{1}{c|}{\multirow{6}[4]{*}{Junction}} & \multicolumn{1}{c|}{\multirow{4}[2]{*}{\makecell[c]{Marginal \\ Predictions}}} & LSTM-ED \cite{Argoverse2019}  & 0.9089  & 2.2090  & 0.6581  & \textbf{0.008817 } & 0.7047  \\
          &       & VectorNet (noGNN) \cite{VectorNet2020}  & 0.8560  & 1.9331  & 0.5753  & 0.012480 & 0.7253  \\
          &       & VectorNet \cite{VectorNet2020}  & 0.7130  & 1.7041  & 0.4538 & 0.011643  & 0.7368  \\
          &       & TNT \cite{TNT2020}   & 0.8180  & 1.8980  & 0.4625  & 0.012556 & 0.7161  \\
\cmidrule{2-8}          & \multicolumn{1}{c|}{\multirow{2}[2]{*}{\makecell[c]{Ablations}}} & CGTP (wo interactive loss) & 0.7022  & 1.6974  & 0.4308  & 0.011217 & 0.7575  \\
          &       & CGTP (w interactive loss) & \textbf{0.6454} & \textbf{1.5481} & \textbf{0.3736} & 0.010379 & \textbf{0.7639}   \\
    \midrule
    \midrule
    \multirow{6}[4]{*}{Non-Junction} & \multicolumn{1}{c|}{\multirow{4}[2]{*}{\makecell[c]{Marginal \\ Predictions}}} & LSTM-ED \cite{Argoverse2019}  & 0.9337  & 2.1816  & 0.6077  & 0.002002 & 0.7965  \\
          &       & VectorNet (noGNN) \cite{VectorNet2020}  & 0.9318  & 2.0643  & 0.5887  & 0.002486 & 0.8151  \\
          &       & VectorNet \cite{VectorNet2020}  & 0.8933  & 1.9584  & 0.5180  & 0.002518 & 0.8018 \\
          &       & TNT \cite{TNT2020}   & 0.9090  & 2.0577  & 0.5285  & 0.002970 & 0.7153  \\
\cmidrule{2-8}          & \multicolumn{1}{c|}{\multirow{2}[2]{*}{\makecell[c]{Ablations}}} & CGTP (wo interactive loss) & 0.7814  & 1.9149  & 0.4721  & 0.002098 & 0.8341  \\
          &       & CGTP (w interactive loss) & \textbf{0.6209} & \textbf{1.4780} & \textbf{0.3544} & \textbf{0.001793} & \textbf{0.8583} \\
    \bottomrule
    \end{tabular}%
  \label{tab:In-house}%
  	 \vspace{-1em} 
\end{table*}%

 \noindent\textbf{OR}. A single overlap is described by a frame where the bounding boxes of two interacting agents overlap with each other at any future time step in the highest confidence trajectory-pair prediction. The average over all frames constructs the overlap rate. 
 
 Here, we use $\bar{\kappa}$ to represent the index of the predicted trajectory-pair with highest confidence score, and define a single overlap indicator over it as 
  \begin{equation} \label{eq:n5-4}
  \begin{gathered}
  IsOverlap(\cdot) =  \mathbb{1}\left(\sum_{\delta=t+1}^{t+T} IOU\left( b\left(\hat{\boldsymbol{y}}_{A}^{\bar{\kappa},\delta}\right), b\left(\hat{\boldsymbol{y}}_{B}^{\bar{\kappa},\delta}\right)
  \right) > 0\right), 
  \end{gathered}
  \end{equation}
 where $b(\cdot)$ is a function to obtain the bounding box information (length, width and heading) from the predicted state of each interacting agent at any future time step $\delta$. Subsequently, inspired by \cite{TrafficSim2021}, $IOU(\cdot)$ computes the intersection-over-union between two bounding boxes of interacting agents.

 \noindent\textcolor{black}{\textbf{mAP}. Given the confidence score of joint predictions estimated by $\pi\left(\boldsymbol{g}_B^k, \boldsymbol{g}_A^{q} \mid \boldsymbol{C}\right)$, mAP calculates the area under the precision-recall curve, where the definition of MR is employed to determine true positives, false positives, $etc$. }
 
 \noindent\textbf{CR}. The Cut-in Rate is computed as the total number of cut-in frames divided by the total number of safety frames. In this paper, we use the definition of OR above to define safety frames. In addition to satisfying the existence of no overlap between two interacting agents, a cut-in frame is determined by a cut-in indicator:
  \begin{equation} \label{eq:n5-5}
  \begin{gathered}
  IsCutin(\cdot) =  \mathbb{1}\left( lane\left(\hat{\boldsymbol{y}}_{A}^{\bar{\kappa},t+T}\right)=lane\left(\hat{\boldsymbol{y}}_{B}^{\bar{\kappa},t+T}\right)\right)
  \\
  \wedge \mathbb{1}\left( y\left(\hat{\boldsymbol{y}}_{B}^{\bar{\kappa},t+T}\right)>y\left(\hat{\boldsymbol{y}}_{A}^{\bar{\kappa},t+T}\right)\right)
  \end{gathered}
  \end{equation}
  where $lane(\cdot)$ denotes a function to calculate the index of the future lane where each interacting agent locates at the last future time step, and $y(\cdot)$ is responsible to derive the longitudinal coordinate of the endpoints for the predicted trajectory-pair.  
 
 \emph{(3) Implementation Details:} This section introduces the implementation details from the aspects of pre-processing, network architecture design and learning scheme.

 \noindent \textbf{Pre-processing}. Towards each interacting agent, both past and future trajectories are normalized at its own reference frame, with the origin centered around the location at its last observed time step. Further, the other observations, including agent trajectories and future lanes, are accordingly transformed to the reference frame of each interacting agent. For the dynamic information, the heuristic rules are performed to select $O$=14 vehicles in the neighbor as surrounding agents. On the other hand, for the static information, we use the Depth-First-Search algorithm to search $P$=6 potential future lanes which each interacting agent is likely to reach, and every future lane has 200 goal candidates sampled at every 0.5 meters. Obviously, we obtain $K$=1200 fine-grained goal candidates in total to represent diverse uncertainties. If the number of surrounding agents or future lanes is insufficient, the corresponding locations are masked out by zeros.

 \noindent \textbf{Network Architecture Design}. In the context encoding aspect, GoINet extracts the individual-level feature over every node via $L$=3 graph layers.
 Due to the existence of the concatenation operator $f_{cc}(\cdot)$, the number of hidden units at graph layer $l$ is twice as that at graph layer $l$-1, and its initial value $d_h$ is set to 16. 
 Subsequently, the goal distribution and offset prediction of two interacting agents, realized by $f_i^{seg}(\cdot)$ and $f_i^{reg}(\cdot)$, are 3-layer MLPs, with the number of hidden units set to 128. Based on the realization of goal interactive prediction, the proposed CGPNet selects the top $\mathcal{K}$=5 future goal candidates of agent $A$ from its marginal goal distribution, which takes turns to be served as a conditional query to determine the same number of future goal candidates of agent $B$ based on the conditional goal distribution. 
 Further, during the trajectory interactive prediction process, a 2-layer bidirectional GRUs are used both by encoder and decode, the hidden dimensions of which are set to 128. Until now, given $\mathcal{K}^2$=25 goal-pairs, 25 goal-oriented trajectory-pairs are produced jointly by employing the proposed GTFNet. \textcolor{black}{Different from Argoverse and In-house cut-in dataset, we further reduce the size of joint trajectory-pairs to 6 to satisfy the evaluation necessity of Waymo motion prediction benchmark. In our CGTP framework, we filter 6 joint predictions from $\mathcal{K}^2$=25 candidates by using the non-maximum suppression method \cite{TNT2020}.}

 \noindent \textbf{Learning Scheme}. Our proposed CGTP framework is trained on 8 A100 GPUs with the Adam optimizer \cite{Adam2014}. The learning rate is initialized as 5e-3, which is decayed by a factor of 0.5 every 30 epochs. Our model approximately requires 200 epochs to train on average with a batch size of 64.
 
 \begin{table*}[t]
  \centering
  \caption{\textcolor{black}{Comparison with interactive prediction benchmark of WOMD}}
    \begin{tabular}{c|c|ccccc}
    \toprule
    \textbf{Joint Prediction} & \textbf{Methods} & \textbf{minADE $\downarrow$} & \textbf{minFDE $\downarrow$} & \textbf{MR $\downarrow$} & \textbf{OR $\downarrow$} &\textbf{mAP $\uparrow$}\\
    \midrule
    \multicolumn{1}{c|}{\multirow{2}[1]{*}{\makecell[c]{Marginal \\ Predictions}}} & Waymo LSTM Baseline \cite{WaymoDataset2021}   & 2.420  & 6.070  & 0.660 & - & 0.070\\
          & TNT \cite{TNT2020}  & 2.585  & 6.136  & 0.605 & 0.186 & 0.167  \\
    \midrule
    \multicolumn{1}{c|}{\multirow{2}[1]{*}{\makecell[c]{Conditional \\ Predictions}}}
          & ProspectNet \cite{ProspectNet}    & 3.012  & 8.118  & 0.826 & 0.416 & 0.115  \\
          & M2I \cite{sun2022m2i}    & 2.399  & 5.477  & 0.552 & 0.174 & 0.177  \\
    \midrule
    \multicolumn{1}{c|}{\multirow{2}[2]{*}{\makecell[c]{Ablations}}} & CGTP (wo interactive loss) & 2.414  & 5.531  & \textbf{0.551} & 0.173 & 0.179  \\
          & CGTP (w interactive loss) & \textbf{2.371} & \textbf{5.395} & 0.559 &\textbf{0.169} & \textbf{0.180}\\
    \bottomrule
    \multicolumn{7}{l}{{$^{\star}$mAP is the official ranking metric.}}
    
    \end{tabular}%

  \label{tab:Waymo}%
\end{table*}%



 \subsection{Quantitative Results}
\noindent\textbf{\textcolor{black}{Comparisons on Argoverse and In-house dataset.}}  We first evaluate the performance of our CGTP framework with the existing mainstream marginal prediction approaches on Argoverse and In-house cut-in dataset. In this paper, we extend LSTM-based encoder-decoder (LSTM-ED) \cite{Argoverse2019}, VectorNet \cite{VectorNet2020} and TNT \cite{TNT2020} to the joint prediction task, producing the marginal predictions for both two agents, without considering their future interactions. Among them, LSTM and VectorNet predict trajectory-pairs via 
pure regression method.
Especially for VectorNet, to validate the effectiveness of message passing in a graph, we provide a variant of VectorNet, named VectorNet(noGNN), whose context representations are purely captured by MLP and Max-pooling. \textcolor{black}{On the other hand, we compare our CGTP framework with the goal-oriented trajectory prediction model TNT to further verify the significance of the proposed goal interactive predictor in our model.} As shown in Table~\ref{tab:Argoverse} and  \ref{tab:In-house}, our proposed model outperforms all marginal approaches on Argoverse and In-house cut-in dataset by a large margin in all distance error metrics (minADE, minFDE and MR). More specifically, for the In-house non-junction environment, the comparative results show that the proposed CGTP framework significantly outperforms the TNT model with 32.9$\%$ reduction in MR. This great enhancement can be attributed to the accurate estimation of joint distribution over future goals by relying on the goal interactive prediction stage.
Also note that VectorNet achieves significant improvements against VectorNet(noGNN), demonstrating that GNN can aggregate interactive features from context information via message passing.


 In terms of interactive metrics like CR and OR, compared with the marginal prediction methods, our CGTP framework can achieve on par or better performance on the In-house cut-in dataset, as shown in Table~\ref{tab:In-house}.
 Specifically in the non-junction environment, the conditional model trained with our proposed framework beats the marginal model trained with TNT, achieving 39.6$\%$ relative reduction on OR and 20.0$\%$ relative gain on CR. Unlike TNT, the proposed CGTP framework models the future interactions at the goal-based level,
 which is capable of learning the joint distribution of  cut-in interactive behaviors. In reverse, the TNT-based marginal model assumes the goal predictions of two interacting agents are independent of each other, which hardly generates reasonable cut-in trajectory-pairs in some scenarios with complex future interactions. \textcolor{black}{Towards the junction environment, our CGTP framework greatly outperforms LSTM-ED in CR, while the performance of OR is slightly worse than it. Such experimental phenomenon results from the poor imitation ability of the simple regression-based marginal model, which may output some 
inaccurate behaviors far away from ground truth cut-in behaviors yet with a safety guarantee between two interacting agents, leading to the illusion of a lower collision rate. }
\noindent \textcolor{black}{\textbf{Comparisons on interactive prediction benchmark of WOMD.} In Table \ref{tab:Waymo}, we compare our model both with marginal and conditional prediction methods. On one hand, the marginal approaches include Waymo LSTM Baseline \cite{WaymoDataset2021} and TNT \cite{TNT2020}, where the former is the official baseline provided by the benchmark and the latter is a typical goal-oriented prediction method that also served as a comparison model for the two datasets above. On the other hand, we take ProspectNet \cite{ProspectNet} and M2I \cite{sun2022m2i} as conditional comparative approaches, which are the state-of-the-art models on the WOMD benchmark of the interactive prediction task. Such conditional models commonly build conditional dependencies on the explicit overall future trajectories while differing in the process of context encoding. In detail, ProspectNet leverages attention to aggregate vectorized features both spatially and temporally, 
while M2I learns features both from rasterized and vectorized representations. In addition, the unique novelty of M2I lies in the proposal of relation predictor, which infers the influencer-reactor relations of two interacting agents and then leverage marginal and conditional trajectory predictor in turn to generate the joint trajectory-pairs. In our work, we adopt the relation predictor of M2I to determine the agent $A$ and $B$ before training and validation, yet focus on the goal interactive prediction in a combined  format of marginal and conditional methods.}

 \begin{figure*}[htbp]
	\centering
	\begin{center}
		\scriptsize
        \includegraphics*[width=0.85\textwidth]{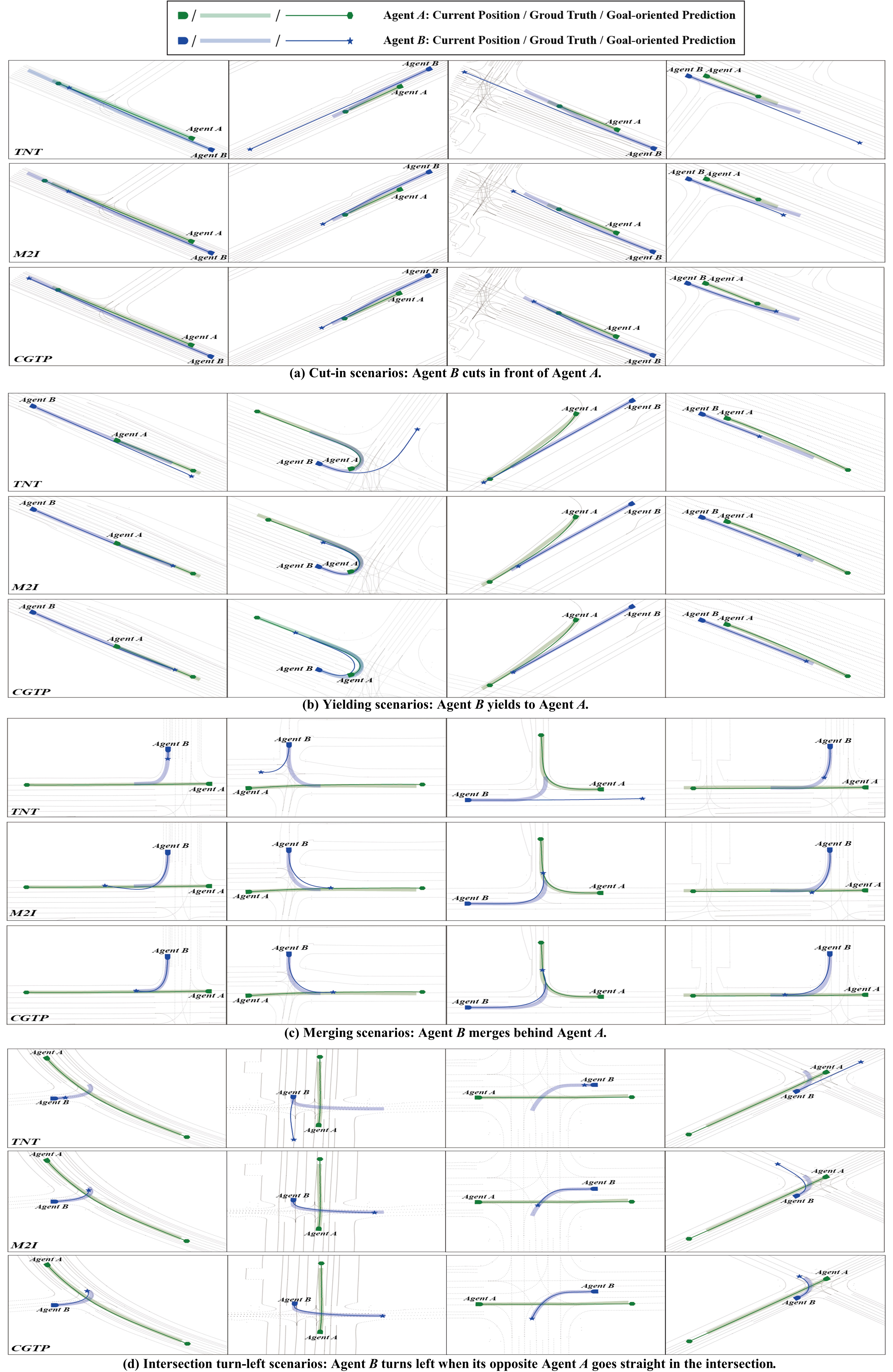}
		\caption{\textcolor{black}{Qualitative examples from TNT, M2I, and our CGTP framework in four classes of pairwise interactive scenarios, including (a) cut-in, (b) yielding, (c) merging, and (d) intersection left-turn. Each pairwise interactive scenario is demonstrated by a group of examples. Compared with TNT (upper row) and M2I (medium row), our CGTP framework (lower row) accounts for future interactions at the goal-based level and achieves better prediction accuracy and scene compliance.
		}  }
		\label{fig:Future-Interaction}
	\end{center}
	 \vspace{-1.3em} 
\end{figure*}

\textcolor{black}{Comparison results demonstrate our model CGTP outperforms Waymo LSTM Baseline in terms of all metrics. Similar to the observations from the two datasets above, the performance of our CGTP framework is superior to the goal-oriented trajectory prediction method TNT.
\textcolor{black}{When compared to the conditional model ProspectNet, our CGTP framework improves mAP by 56.52$\%$, meaning that the combined design of CGPNet and GTFNet is capable of learning more accurate joint distribution in the future.} 
\textcolor{black}{We further validate the effectiveness of future interactive modeling between the sparse explicit future goals by comparing the performance of joint predictions using our model and state-of-the-art model M2I, with the latter considering future interactions between the redundant explicit future trajectories instead, and achieve 2.87$\%$ reduction in OR and 1.69$\%$ gain in mAP, the official ranking metric.} 
}

\noindent\textcolor{black}{\textbf{Comparisons on ablation study.}} We conduct ablation studies to analyze the contribution of interactive loss in the proposed CGTP framework. As shown in Table \ref{tab:Argoverse} - \ref{tab:In-house}, our current CGTP framework achieves on par or better results in all metrics by adding a novel interactive loss. The improvements indicate that our model with the interactive loss can obtain the high-quality goal-pair estimated by the learned joint distribution, and then produces the scene-compliant goal-oriented trajectory-pair most closely matching the ground truth. \textcolor{black}{Towards WOMD, the most likely trajectory-pair learned by the interactive loss can characterize a more reasonable interactive behavior in the future, improving mAP metric by 0.56$\%$ gain.} Similar observations are also presented in the interactive metrics of the In-house cut-in dataset.
 More specifically, in the junction environment, OR and CR is 7.5$\%$ and 0.8$\%$ better compared to our model without interactive loss. In the non-junction environment, our model with interactive loss is 14.5$\%$/2.9$\%$ better in OR/CR compared to the one without interactive loss. 

\subsection{Qualitative Results}

\textcolor{black}{In Fig.~\ref{fig:Future-Interaction}, we present four classes of challenging pairwise interactive scenarios in WOMD, including cut-in, yielding, merging and intersection left-turn, and visualize the most likely trajectory-pair from the goal-oriented trajectory prediction method TNT, the SOTA method M2I,  and our CGTP framework, respectively. In Fig.~\ref{fig:Future-Interaction}.(a), a group of examples depicts a pairwise interactive scenario where agent $B$ is cutting in front of agent $A$. The goal-oriented trajectory prediction model TNT fails to capture the interaction and predict overlapping trajectories, as shown in the first column of Fig.~\ref{fig:Future-Interaction}.(a). In spite of no overlap exhibited in the remaining cut-in examples, TNT also results in less accurate predictions that mismatch the ground truth interactive behaviors. Also, M2I hardly captures accurate cut-in aggressive interactive behaviors by considering the overall predicted trajectory of agent $A$. Instead, our CGTP framework is sensitively aware of the underlying interaction between future goals of two interacting agents, and predicts an accurate endpoint of agent $B$ conditioned on the predicted endpoint of agent $A$, and then outputs a scene-compliant goal-oriented trajectory-pair given an accurate cut-in goal-pair prediction. Different from Fig.~\ref{fig:Future-Interaction}.(a), Fig.~\ref{fig:Future-Interaction}.(d) provides a set of examples where agent $B$ turns left when its opposite agent $A$ goes straight at the intersection, which represents a more challenging pairwise interactive scenario. In each example, our CGTP framework successfully improves prediction accuracy and scene compliance, while TNT predicts trajectories far away from the ground truth without considering the future interaction between two agents.}

\section {Conclusion}\label{sec:Conclusion}
In this paper, we propose a novel CGTP framework for interactive behavior prediction of two agents. We build the hierarchical representations of fine-grained future goals, and focus on the goal interactive prediction stage by a combined form of marginal and conditional goal predictors, where we predict the future goals of agent $A$ via marginal goal predictor and then perform future goal prediction of agent $B$ conditioned on per marginal prediction. Once the goal-pairs of two interacting agents are determined, a trajectory interactive prediction module is designed to generate the goal-oriented trajectory-pairs in a step-wise rollout manner. \textcolor{black}{The experimental results  conducted on Argoverse motion forecasting dataset, In-house cut-in dataset and Waymo open motion dataset show the superiority of our proposed method in prediction accuracy and scene compliance. As future work, the joint prediction for more interacting agents with low computational burden is an interesting and important frontier field.}
\ifCLASSOPTIONcaptionsoff
  \newpage
\fi





\bibliographystyle{IEEEtran}
\bibliography{IEEEabrv,Bibliography}

\vfill


\end{document}